\documentclass[10pt,twocolumn,letterpaper]{article}

\usepackage{cvpr}              

\usepackage[table]{xcolor}

\usepackage{graphicx}
\usepackage{amsmath}
\usepackage{amssymb}
\usepackage{booktabs}

\usepackage{tikz}
\usepackage{comment}
\usepackage{amsmath,amssymb} 
\usepackage{bbm}
\usepackage{booktabs}
\usepackage[colorlinks, pagebackref]{hyperref}
\usepackage{graphicx}
\usepackage{bm}
\usepackage{multirow}

\usepackage{cases}

\usepackage{color}
\usepackage{ulem}

\usepackage{algorithm}
\usepackage{algorithmic}
\usepackage{varwidth}

\usepackage{hyperref}
\usepackage{url}
\usepackage{times}
\usepackage{epsfig}
\usepackage{graphicx}
\usepackage{amsmath}
\usepackage{amssymb}
\usepackage{amsthm}
\usepackage{multirow}
\usepackage{soul}
\usepackage{footnote}
\usepackage{tablefootnote}
\usepackage{caption}
\usepackage{subcaption}
\usepackage{mathtools}

\usepackage[capitalize]{cleveref}
\crefname{section}{Sec.}{Secs.}
\Crefname{section}{Section}{Sections}
\Crefname{table}{Table}{Tables}
\crefname{table}{Tab.}{Tabs.}

\def\cvprPaperID{1383} 
\def\confName{CVPR}
\def\confYear{2023}

\definecolor{top1}{RGB}{255,179,179}
\definecolor{top2}{RGB}{255,217,179}
\definecolor{top3}{RGB}{255,255,179}

\usepackage{xcolor,pifont}
\newcommand*\colourcheck[1]{%
  \expandafter\newcommand\csname #1check\endcsname{\textcolor{#1}{\ding{52}}}%
}
\colourcheck{blue}
\colourcheck{green}
\colourcheck{red}
\usepackage{makecell}

\begin{document}

\title{StegaNeRF: Embedding Invisible Information within Neural Radiance Fields}

\author{
  \textbf{Chenxin Li\textsuperscript{1}\thanks{Equal contribution}, Brandon Y. Feng\textsuperscript{2$\ast$}, Zhiwen Fan\textsuperscript{3$\ast$}, Panwang Pan\textsuperscript{4}, Zhangyang Wang\textsuperscript{3}}\\
  {\textsuperscript{1}Hong Kong Polytechnic University}, \, {\textsuperscript{2}University of Maryland},\\ {\textsuperscript{3}University of Texas at Austin}, \,{\textsuperscript{4}ByteDance}\\
}

\maketitle

\begin{abstract}
Recent advances in neural rendering imply a future of widespread visual data distributions through sharing NeRF model weights.
However, while common visual data (images and videos) have standard approaches to embed ownership or copyright information explicitly or subtly, the problem remains unexplored for the emerging NeRF format.
We present {StegaNeRF}, a method for steganographic information embedding in NeRF renderings.
We design an optimization framework allowing accurate hidden information extractions from images rendered by NeRF, while preserving its original visual quality.
We perform experimental evaluations of our method under several potential deployment scenarios, and we further discuss the insights discovered through our analysis.
StegaNeRF signifies an initial exploration into the novel problem of instilling customizable, imperceptible, and recoverable information to NeRF renderings, with minimal impact to rendered images.
Project page: \url{https://xggnet.github.io/StegaNeRF/}.

\end{abstract}


\section{Introduction}
\label{sec:intro}



Implicit neural representation (INR) is an emerging concept where the network describes the data through its weights~\cite{mescheder2019occupancy, park2019deepsdf, sitzmann2020implicit, tancik2020fourier, mildenhall2021nerf}.
After training, the INR weights can then be used for content distribution, streaming, and even downstream inference tasks, all without sending or storing the original data.
Arguably the most prominent INR is Neural Radiance Fields (NeRF)~\cite{mildenhall2021nerf}, where a network learns a continuous function mapping spatial coordinates to density and color.
Due to its lightweight size and superb quality, NeRF has immense potential for 3D content representation in future vision and graphics applications.

While there is a plethora of work dedicated towards better quality~\cite{barron2021mip,tancik2022block, wang2021neus, Zhang2021NeRFactorNF}, faster rendering~\cite{muller2022instant,sun2022direct,wang2022fourier,schwarz2022voxgraf,yu2021plenoxels}, and sparse view reconstruction~\cite{niemeyer2022regnerf,zhang2022ray,chen2022geoaug,xu2022sinnerf,yu2021pixelnerf,chen2021mvsnerf}, in this paper, we look beyond the horizon and explore a new question: Can we achieve {steganography} with NeRF?

\begin{figure}[!t]   
	\centering	   
\includegraphics[width=\linewidth]{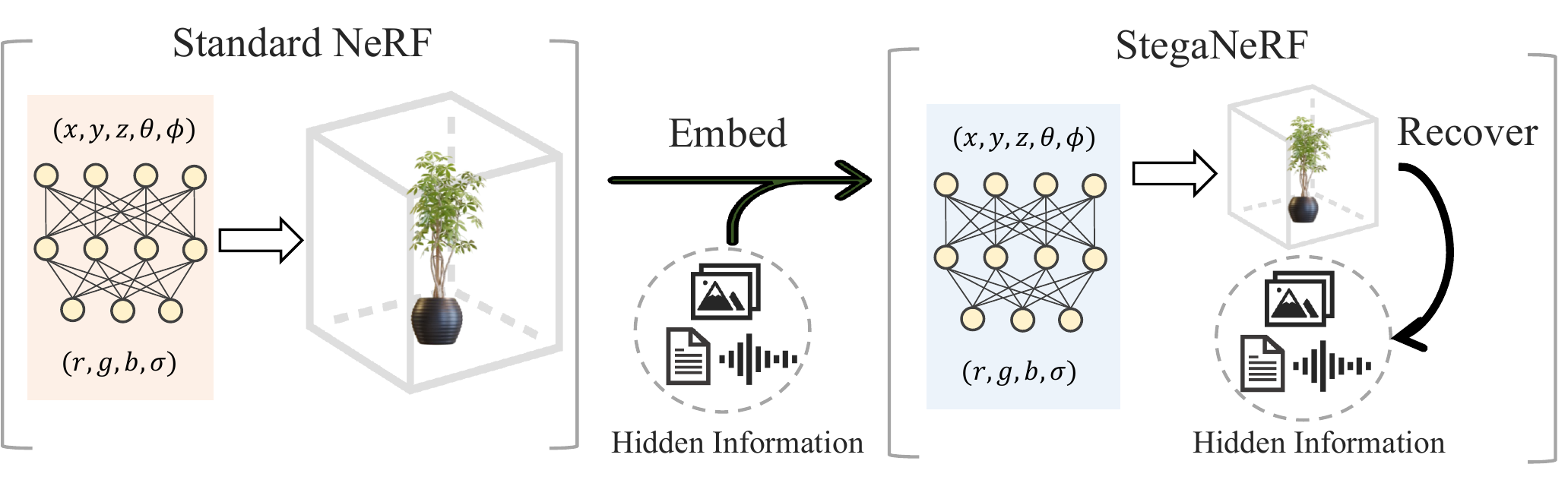}  
\vspace{-1em}
	\caption{We introduce the new problem of NeRF steganography: hiding information in NeRF renderings. Our proposed framework, StegaNeRF, can embed and recover customized hidden information while preserving the original NeRF rendering quality.}
  \label{FIG:ERASER}   
\vspace{-15pt}
\end{figure}

Established digital steganography method~\cite{cox2007digital}
focus on embedding hidden messages in 2D images.
The recent growth of deep learning and social media platforms further gives rise to many practical use cases of image steganography.
As countless images and videos are shared online and even used to train deep learning models, 2D steganography methods~\cite{baluja2017hiding, baluja2019hiding} allow users and data providers to protect copyright, embed ownership, and prevent content abuse.

Now, with the ongoing advances in 3D representations powered by NeRF, we envision a future where people frequently share their captured 3D content online
just as they are currently sharing 2D images and videos online.
Moreover, we are curious to explore the following research questions:
\ding{202} Injecting information into 2D images for copyright or ownership identification is common, but can we preserve such information when people share and render 3D scenes through NeRF? \ding{203} NeRF can represent large-scale real-world scenes with training images taken by different people, but can we preserve these multiple source identities in the NeRF renderings to reflect the collaborative efforts required to reconstruct these 3D scenes? \ding{204} Common image steganography methods embed either a hidden image or a message string into a given image, but can we allow different modalities of the hidden signal in NeRF steganography?

Driven by these questions, we formulate a framework to embed customizable, imperceptible, and recoverable information in NeRF renderings without sacrificing the visual quality.
Fig.\,\ref{FIG:ERASER} presents an overview of our proposed framework, dubbed {StegaNeRF}.
Unlike the traditional image steganography that embeds hidden signals only into a specific source image, we wish to recover the same intended hidden signal from NeRF rendered at arbitrary viewpoints.

Despite many established works on 2D steganography and hidden watermarks for image and video~\cite{chang2003finding, baluja2017hiding, baluja2019hiding,jing2021hinet, lu2021large,tancik2020stegastamp},
%
naively applying 2D steganography on the NeRF training images is not practical, since the embedded information easily gets lost in the actual NeRF renderings.
In contrast, our framework enables reliable extraction of the hidden signal from NeRF renderings. 
During NeRF training, we jointly optimize a detector network to extract hidden information from the NeRF renderings.
To minimize the negative impact on the visual quality of NeRF, we identify weights with low impact on rendering and introduce a gradient masking strategy to steer the hidden steganographic information towards those low-importance weights.
%
Extensive experimental results validate StegaNeRF balances between the rendering quality of novel views and the high-fidelity transmission of the concealed information.

StegaNeRF presents the first exploration of hiding information in NeRF models for ownership identification, and our contributions can be summarized as follows: 
\begin{itemize}
\item{We introduce the new problem of NeRF steganography and present the first effort to embed customizable, imperceptible, and recoverable information in NeRF.}
\item{We propose an adaptive gradient masking strategy to steer the injected hidden information towards the less important NeRF weights, balancing the objectives of steganography and rendering quality.}
\item{We empirically validate the proposed framework on a diverse set of 3D scenes with different camera layouts and scene characteristics, obtaining high recovery accuracy without sacrificing rendering quality.}
\item {We explore various scenarios applying StegaNeRF for ownership identification, with the additional support of {multiple identities} and multi-modal signals.}
%
\end{itemize}

\section{Related Work}


\paragraph{Neural Radiance Fields}

\begin{figure*}[!t]   
	\centering	   \includegraphics[width=1.0\linewidth]{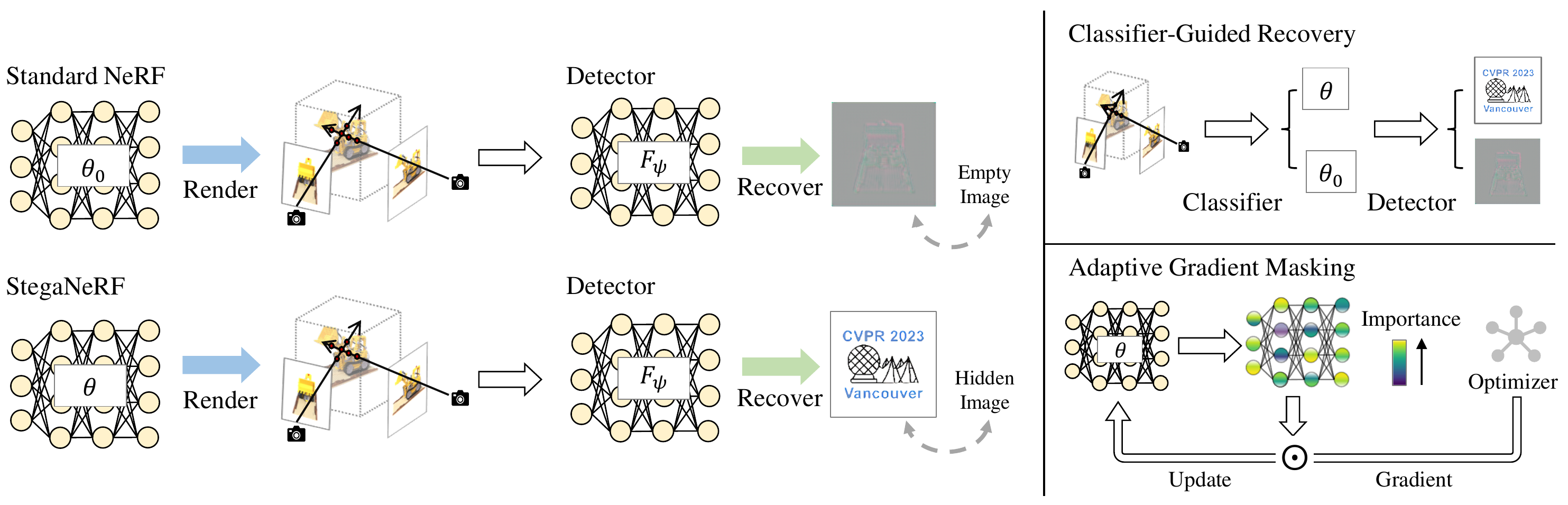}     
 	\vspace{-15pt}
	\caption{ {StegaNeRF training overview. At the first stage, we optimize $\boldsymbol{\theta_{0}}$ with standard NeRF training. At the second stage, we initialize $\boldsymbol{\theta}$ with $\boldsymbol{\theta_{0}}$ and optimize for the steganography objectives. We train the detector $\boldsymbol{F_{\psi}}$ to recover hidden information from StegaNeRF renderings and no hidden information from original NeRF renderings. We introduce {\it Classifier Guided Recovery} to improve the accuracy of recovered information, and {\it Adaptive Gradient Masking} to balance between steganography ability and rendering visual quality.
	}
 }
  \label{FIG:METHOD}    
\end{figure*} 

The success of NeRF~\cite{mildenhall2021nerf} sparks a promising trend of improving highly photo-realistic view synthesis with regard to its rendering quality.
A wide range of techniques have been incorporated into NeRF, including ray re-parameterizations~\cite{barron2021mip,Barron2022MipNeRF3U}, explicit spatial data structures~\cite{yu2021plenoctrees, fridovich2022plenoxels,liu2020neural,hu2022efficientnerf,sun2022direct,muller2022instant}, caching and distillation~\cite{wang2022r2l, garbin2021fastnerf,hedman2021baking,reiser2021kilonerf}, ray-based representations~\cite{attal2021learning, Sitzmann2021LightFN, Feng2021SIGNETEN}, geometric primitives~\cite{Lin2022NeurMiPsNM, Lombardi2021MixtureOV}, large-scale scenes~\cite{zhang2020nerf++,martin2021nerf,xiangli2022bungeenerf,tancik2022block}, and dynamic settings~\cite{park2021nerfies,pumarola2021d,gafni2021dynamic,li2022neural,xian2021space}.  
Unlike prior works that make NeRF efficient or effective, this paper explores the uncharted problem of embedding information in NeRF renderings, with critical implications for copyright protection and ownership preservation.
As early-stage NeRF-based products already become available~\cite{lumalabs, polycam}, we believe more activities based on 3D NeRFs will quickly emerge, and now is the right time to open up the problem of NeRF steganography. 

\paragraph{Image Steganography}
Steganography hides intended signals as invisible watermarks (\textit{e.g.}, hyperlinks, images) within the cover media called carriers (\textit{e.g.}, images, video and audio)~\cite{cox2007digital,kessler2011overview}.
%
Classical methods focus on seeking and altering the least significant bits (LSBs)~\cite{fridrich2001detecting,tamimi2013hiding,wolfgang1996watermark,pevny2010using} and transforming domain techniques~\cite{chandramouli2003image,wang2004cyber,bi2007robust,qian2015deep}.
Prior research also uses deep neural networks for steganography~\cite{hayes2017generating,tang2017automatic, baluja2017hiding,baluja2019hiding}. 
Among them, DeepStega~\cite{baluja2017hiding} conceals a hidden image within an equal-sized carrier image.
Subsequent works~\cite{jing2021hinet,lu2021large} use invertible networks to improve the performance of deep image steganography.
Another line of work conceals information in other carrier media like 
audio~\cite{cui2020multi, geleta2022pixinwav,yang2019hiding}
and video~\cite{weng2019high,luo2021dvmark,sadek2015video}.
%
%
The above advances all play a critical part in the era when traditional media formats like images and videos are dominant.
However, as MLP-based neural representations of 3D scenes are gaining momentum to become a major format of visual data, extending steganography to NeRF is bound to become an important problem in the upcoming future.

\paragraph{Lifting Steganography to 3D} 
Prior to NeRF, meshes are commonly used to represent 3D shapes.
Early pioneers apply steganography 3D meshes~\cite{benedens1999geometry, praun1999robust} for copyright protection when meshes are exchanged and edited.
More recent work~\cite{wu2021embedding} has also explored embedding multi-plane images within a JPEG-compressed image. Their problem can be regarded as a special case of 2D steganography, hiding multiple 2D images inside a single 2D image.
In contrast, we try to hide a natural image into a 3D scene representation (NeRF), fundamentally differing from these prior methods where 2D images act as the carrier of hidden information.

%






\section{Method}
The goal of StegaNeRF is to inject customized (steganographic) information into the NeRF weights with imperceptible visual changes when rendering.
Given a NeRF with weights ${\boldsymbol{\theta_0}}$ and the information $I$ to be injected, when we render with ${\boldsymbol{\theta}}$ on any viewpoint, we hope the injected $I$ can be recovered by a detector $F_{{\psi}}$ with learnable weights $\psi$.

A seemingly obvious solution is to use prior image steganography methods to 1) inject $I$ on training images, 2) train NeRF on those images with embedded information, 3) apply their provided $F_{{\psi}}$ to extract $I$ from NeRF-rendered images. This approach has been successfully applied for GAN fingerprinting~\cite{yu2021artificial}. However, it fails in the NeRF context (see Fig.~\ref{FIG:CHALLENGE}), where the subtle changes induced by prior steganography methods easily get smoothed out, inhibiting the detector from identifying the subtle patterns necessary for information recovery.

Such shortcomings of off-the-shelf steganographic methods are not surprising since they are developed for the traditional setting where 2D images are the ultimate form of visual information.
In contrast, our problem setting involves the new concept of INR as the underlying representation, and 2D images are just the final output rendered by NeRF.

\subsection{Two-Stage Optimization}
Recognizing the formulation difference due to the emergence of implicit representation with NeRF, we move away from the traditional 2D image-based pipeline that trains an encoder network to inject subtle changes to the given 2D images.
Instead, we incorporate the steganography objective into the gradient-based learning process of NeRF.

We re-design the training as a two-stage optimization.
The first stage is the original NeRF training procedure, involving the standard photometric loss between the rendered and ground truth pixels to guarantee the visual quality. 
After finishing the weights update $\boldsymbol{\theta_{0}}$ at the first stage,
we dedicate the second stage to obtain the final NeRF weights $\boldsymbol{\theta}$ containing steganographic information.
We introduce several techniques to achieve robust information recovery with imperceptible impact on the rendered images. The training workflow is depicted in Fig.~\ref{FIG:METHOD} and Alg.\,\ref{ALG1}. 

\subsection{Information Recovery}\label{SEC:1}
Let $P$ denote the camera pose at which we render an image from a NeRF network.
We want to recover $I$ from the rendered image $\boldsymbol{\theta}(P)$.
Importantly, we also want to avoid false positives on images $\boldsymbol{\theta}_{0}(P)$ rendered by the original network without steganography ability, even if the images rendered by $\boldsymbol{\theta}_{0}$ and $\boldsymbol{\theta}$ look visually identical.
Therefore, we optimize $\boldsymbol{\theta}$ to minimize the following contrastive loss terms:

\begin{equation}
\mathcal{L}^+_{{dec}} = | F_{\psi}(\boldsymbol{\theta}(P)) - I |, \quad
\mathcal{L}^-_{{dec}} =  | F_{\psi}(\boldsymbol{\theta}_{0}(P)) - \varnothing |,
\label{EQ:DEC1}
\end{equation}
where $\varnothing$ is an empty image with all zeros.

\begin{figure}[!t]   
	\centering	   
	\includegraphics[width=\linewidth]{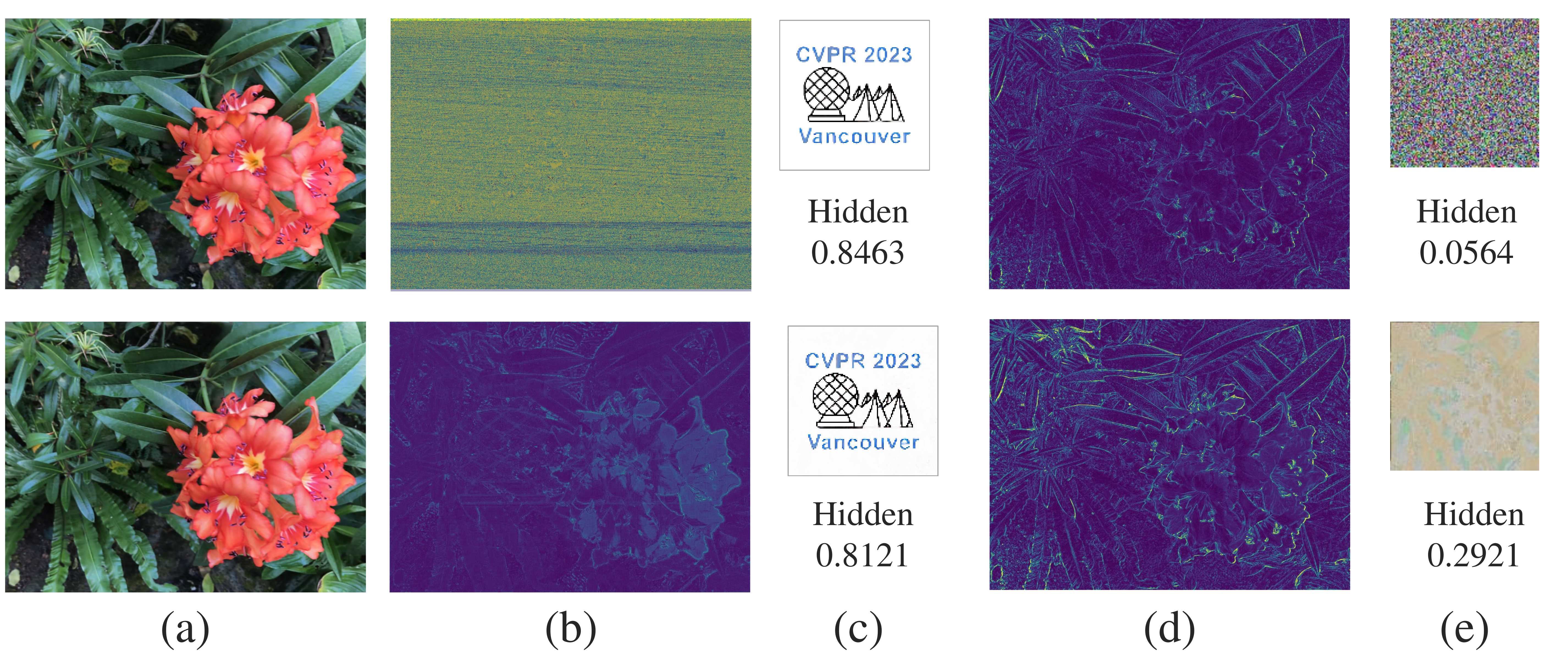}         
	\caption{
    Results of applying prior 2D steganography methods, LSB~\cite{chang2003finding} (top) and DeepStega~\cite{baluja2017hiding} (bottom).
    From left to right, we show (a) training image after applying 2D steganography, (b) residual error of (a) over ground truth, (c) hidden image recovered from (a), (d) residual error of NeRF rendering at the pose of (a), and (e) hidden image recovered from (d). 
    Prior 2D steganography methods fail in the NeRF context since the hidden information injected in training images mostly disappear in NeRF renderings.
}
  \label{FIG:CHALLENGE}    
\end{figure} 

\paragraph{Classifier-Guided Recovery}
The detector $F_{{\psi}}$ is easily implemented as a U-Net to decode $I$ as the form of 2D images, but accurately recovering all the details in $I$ might be challenging.
Therefore, we additionally train a classifier network $F_{{\psi}}^{c}$  to solve the easier task of classifying whether the given NeRF rendering contains hidden information.
$F^c_{\psi}$ is optimized by the following cross-entropy loss:
\begin{equation}
\mathcal{L}^c_{dec} = -\log \big( F^c_{\psi}(\boldsymbol{\theta}(P)) \big) - \log \big(1- F^c_{\psi}(\boldsymbol{\theta}_{0}(P)) \big).
\label{EQ:DEC2}
\end{equation}
We then use its prediction to guide the process of decoding pixel-wise information by adding the classification output as input to $F_{{\psi}}$, such that $F_{{\psi}}(x) = F_{\psi}\big(x ,F^c_{\psi}(x) \big)$.

%

Although the above discussion focuses on hiding images, our framework can be easily extended to embed other modalities like strings, text, or even audio, all of which may be represented as 1D vectors.
We can simply modify the architecture of $F_{\psi}$ to have a 1D prediction branch.

\begin{algorithm}[!t]
  \caption{Train StegaNeRF on a single scene}
  \label{ALG1}
  \textbf{Data}: Training images $\{Y_{i}\}$ with poses $\{P_{i}\}$, hidden information $I$, learning rate $\eta=[\eta_0,\eta_1]$\\
  \textbf{Output}: Steganographic NeRF ${\boldsymbol{\theta}}$ and detector $F_\psi$\\
  Initialize NeRF ${\boldsymbol{\theta}_0}$ and detector network $F_\psi$\\
  Optimize ${\boldsymbol{\theta}_0}$ on $\{Y_{i}, P_{i}\}$ with standard NeRF training\\
  Compute mask $\boldsymbol{m}$ for ${\boldsymbol{\theta_0}}$ as in Eq.~(\ref{EQ:REG1})\\
  \textbf{for} each training iteration $t$ ~\textbf{do}
\begin{algorithmic}
    \STATE Randomly sample a training pose $P_{i}$ and render $\boldsymbol{\theta}(P_{i})$ \\
    \STATE Compute stega. losses  $\mathcal{L}^{c}_{dec}$, $\mathcal{L}^{+}_{dec}$, $\mathcal{L}^{-}_{dec}$ as Eq.~(\ref{EQ:DEC1}),~(\ref{EQ:DEC2})\\
    \STATE Compute standard loss $\mathcal{L}_{rgb}$ in Eq.~(\ref{EQ:REG3})\\
    \STATE Combine total loss $\mathcal{L}$ as in Eq.~\eqref{EQ:OVERALL}\\
    \STATE Update $\boldsymbol{\theta}$ with  $\eta_{0}\cdot(\frac{\partial \mathcal{L}}{{\partial \boldsymbol\theta}} \odot \boldsymbol{m})$ and $F_\psi$ with $\eta_{1}\cdot\frac{\partial \mathcal{L}}{{\partial \psi }}$
\end{algorithmic}
  \textbf{end for}
\end{algorithm}

\subsection{Preserving Perceptual Identity}\label{SEC:2}
Since we want to hide information without affecting the visual perception of the rendered output, an intuitive regularization is to penalize how much $\boldsymbol{\theta}$ deviates from $\boldsymbol{\theta_{0}}$.
However, we find that naively summing the deviation penalty across all weights makes it difficult for the NeRF network to adjust its weights for the steganographic objective.
Instead, motivated by the fact that INR weights are not equally important and exhibit strong sparsity~\cite{lee2021meta,yuce2022structured}, we propose an adaptive gradient masking strategy to encode the hidden information on specific weights' groups.

Formally, given the initial weights ${\boldsymbol{\theta_{0}}}\in\mathbb{R}^N$, we obtain the importance of weights $\boldsymbol{w}$ and a mask $\boldsymbol{m} \in \mathbb{R}^N$ as
\begin{equation}
\boldsymbol{w} = \left|\boldsymbol{\theta_{0}} \right |^\alpha, \quad 
\boldsymbol{m} =  \frac{\boldsymbol{w}^{-1}}{\sum_{i}^{N} \boldsymbol{w}_{i}^{-1}},
\label{EQ:REG1}
\end{equation}
where $\alpha>0$ controls the relative distribution of importance across the weights.
We mask the gradient as $\frac{\partial \mathcal{L}}{{\partial \boldsymbol\theta }} \odot \boldsymbol{m}$ when optimizing $\boldsymbol\theta$ based on the total loss $\mathcal{L}$ in the second stage, where $\odot$ is a Hadamard product.
Effectively, more significant weights are ``masked out'' to minimize the impact of steganographic learning on the rendered visual quality.


We retain the photometric error of the vanilla NeRF~\cite{mildenhall2021nerf} formulation in the steganography learning to prevent NeRF from deviating from its rendered visual signal fidelity:
\begin{equation}
\mathcal{L}_{rgb} = | \boldsymbol{\theta}(P) - \boldsymbol{\theta}_{0}(P) |,
\label{EQ:REG3}
\end{equation}

The overall training loss at the second stage can be formulated as follows:
\begin{equation}
\mathcal{L} = \lambda_0 \mathcal{L}^c_{dec} + \lambda_1\mathcal{L}^+_{{dec}} + \lambda_2\mathcal{L}^-_{{dec}} +  \lambda_3 \mathcal{L}_{rgb}.
\label{EQ:OVERALL}
\end{equation}

\begin{algorithm}[!t]
\caption{Typical usage scenario of StegaNeRF}
\label{ALG2}
\begin{algorithmic}[1]
\STATE Alice captures some images of a 3D scene \\
\STATE Alice trains a StegaNeRF to hide a personalized image \\
\STATE Alice shares the model ${\boldsymbol{\theta}}$ online for other people to enjoy and explore the 3D scene themselves \\
\STATE Bob grabs the model ${\boldsymbol{\theta}}$ and reposts it with his own account without crediting Alice or asking for permission \\
\STATE Alice sees Bob's post, deploys the detector $F_\psi$, and verifies the owner of ${\boldsymbol{\theta}}$ is Alice, not Bob \\
\STATE Bob takes down the post or gets banned for copyright infringement
\end{algorithmic}
\end{algorithm}

\section{Experiments}
In this section, we present experimental evaluations under several use case scenarios.
We further provide additional analysis on the impact of each proposed technique and the robustness analysis of the overall framework.

\subsection{Implementation Details.}
\textit{Dataset.}
   We use common datasets LLFF~\cite{mildenhall2019local} and NeRF-Synthetic~\cite{mildenhall2021nerf}, with forward scenes \{\textit{flower, fern, fortress, room}\} from LLFF and $360^{\circ}$ scenes \{\textit{lego, drums, chair}\} from  NeRF-Synthetic.
   We further experiment on the \textit{Brandenburg Gate} scene from NeRF-W dataset~\cite{martin2021nerf}, with over 800 views of in-the-wild collected online.   

\textit{Training.}
    On LLFF and NeRF-Synthetic, we adopt Plenoxels~\cite{yu2021plenoxels} as the NeRF backbone architecture for efficiency. 
    On NeRF-W, we use the available PyTorch implementation~\cite{queianchen_nerf}.
    The first stage of training is performed according to the standard recipes of those implementations.
    We then perform the second stage of steganography training for 55 epochs unless otherwise noted.
    On LLFF and NeRF-W, we downsize the training images by 4 times following common practice, and we use the original size on NeRF-Synthetic.
    %
    For hyper-parameters in Eq.~(\ref{EQ:OVERALL}), we set the weight of NeRF reconstruction error $\lambda_3=1$ for all experiments.
    We set $\lambda_0=0.01, \lambda_1=0.5, \lambda_2=0.5$ for all scenes on LLFF dataset, and $\lambda_0=0.1, \lambda_1=1, \lambda_2=1$ for the scenes in NeRF-Synthetic, and $\lambda_0=0.05,\lambda_1=1,\lambda_2=1$ for NeRF-W. 
    For Eq.~(\ref{EQ:REG1}), we set $\alpha=3$ for all scenes.
    We run experiments on one NVIDIA A100 GPU. 

\begin{table}[!t]
  \centering
  \caption{
  Quantitative results of StegaNeRF rendering and hidden information recovery.
  {\it Standard NeRF} is the initial NeRF $\boldsymbol{\theta_{0}}$ with standard training, serving as an upper-bound performance for NeRF rendering.
  %
  Prior 2D steganography fails after NeRF training while StegaNeRF successfully embeds and recovers hidden information with minimal impact on the rendering quality.
  Results are averaged over the selected LLFF and NeRF-Synthetic scenes.
  }
  \resizebox{1.0\linewidth}{!}{
  \begin{tabular}{l|ccc|cc}
 
    \toprule
    \multirow{2}[4]{*}{{Method}} & \multicolumn{3}{c|}{NeRF Rendering} & \multicolumn{2}{c}{Hidden Recovery} \\
\cmidrule{2-6}          & PSNR\,$\uparrow$  & SSIM\,$\uparrow$  & LPIPS\,$\downarrow$ & Acc.\,(\%)\,$\uparrow$ & SSIM\,$\uparrow$ \\
    \midrule
    Standard NeRF & 27.74 & 0.8353  & 0.1408  & 50.00  & N/A  \\
    \midrule
    LSB~\cite{chang2003finding}   & \cellcolor{top1}27.72 & \cellcolor{top1}0.8346   &\cellcolor{top1}0.1420  & N/A     & \cellcolor{top3}0.0132  \\
    DeepStega~\cite{baluja2017hiding} & \cellcolor{top3}26.55 & \cellcolor{top3}0.8213  & \cellcolor{top3}0.1605  & N/A     & \cellcolor{top2}0.2098  \\
    StegaNeRF (Ours) & \cellcolor{top2}27.72 & \cellcolor{top2}0.8340  & \cellcolor{top2}0.1428  & \cellcolor{top1}100.0  & \cellcolor{top1}0.9730  \\
    
    \midrule
    \midrule
    Standard NeRF & 31.13 & 0.9606  & 0.0310  & 50.00  & N/A  \\
    \midrule
    LSB~\cite{chang2003finding}   & \cellcolor{top2}31.12 & \cellcolor{top2}0.9604  & \cellcolor{top2}0.0310  & N/A     & \cellcolor{top3}0.0830  \\
    DeepStega~\cite{baluja2017hiding} & \cellcolor{top1}31.13 & \cellcolor{top1}0.9606  & \cellcolor{top1}0.0313  & N/A     & \cellcolor{top2}0.2440  \\
    StegaNeRF (Ours) & \cellcolor{top3}30.96 & \cellcolor{top3}0.9583  & \cellcolor{top3}0.0290  & \cellcolor{top1}99.72 & \cellcolor{top1}0.9677  \\
    \bottomrule
    \end{tabular}%
    }
  \label{TAB:EXP-1}%
\end{table}%

\begin{figure*}[!t]   
\centering	   
\includegraphics[width=\linewidth]{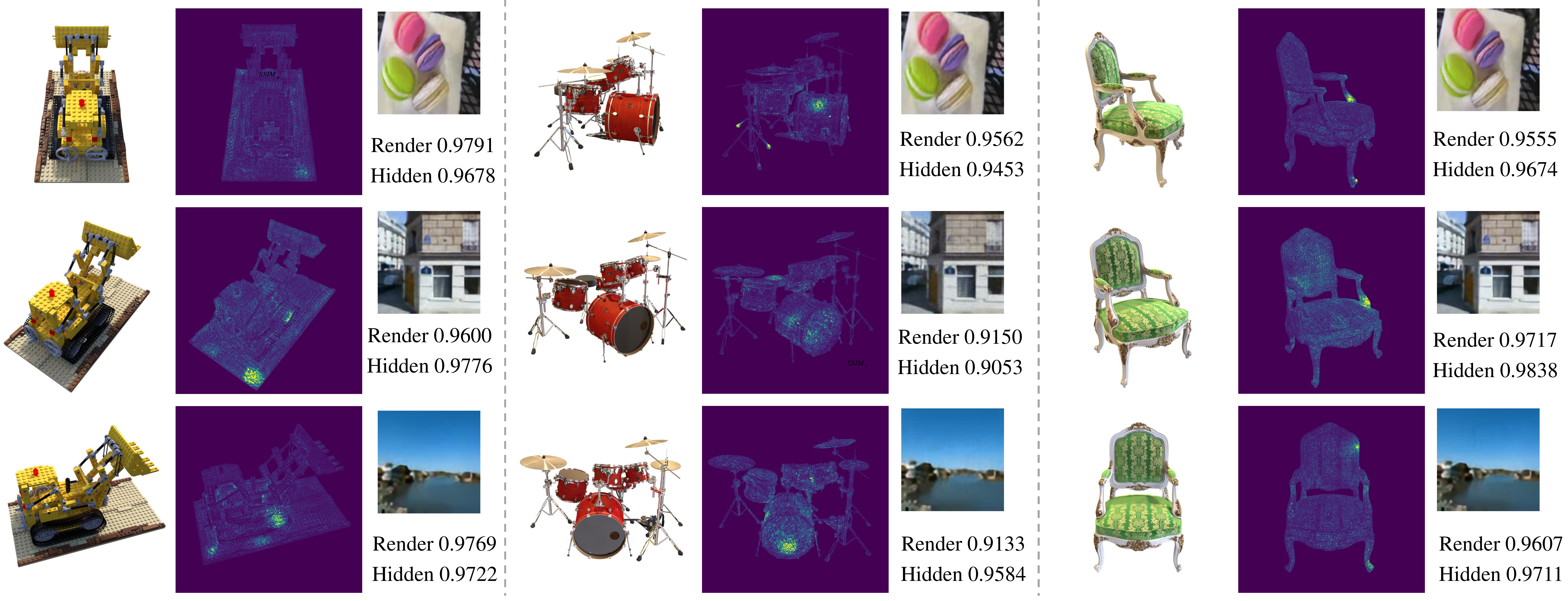}       \vspace{-0.5em}
\caption{Results on the NeRF-Synthetic dataset.
Within each column, we show the StegaNeRF rendering, residual error from the initial NeRF rendering, and the recovered hidden image.
We show the SSIM for the StegaNeRF renderings and the recovered hidden images.
    }
 \label{FIG:EXP-1}    
\end{figure*} 

\textit{Evaluation.}
    We evaluate our system based on a typical authentication scenario shown in Alg.\,\ref{ALG2}.
   %
   We consider the recovery quality of the embedded information, including the metrics of classification accuracy (Acc.) of the classifier, and the structural similarity (SSIM) ~\cite{wang2004image} of the hidden image recovered by the detector.
    We evaluate the final NeRF renderings with PSNR, SSIM and LPIPS~\cite{zhang2018unreasonable}.
    All the metrics are computed on the test set and averaged over all the scenes and embedded images.
    Per-scene details are provided in the supplement.
    %

\subsection{Case I: Embedding in a Single Scene.} \label{SEC:EXP-1}
   We first explore the preliminary case of ownership identification on a specific NeRF scene. We select random images from ImageNet~\cite{krizhevsky2017imagenet} as the hidden information to be injected to NeRF renderings. 
    \paragraph{Failure of 2D Baseline} 
    Due to the lack of prior study on NeRF steganography, we consider a baseline from 2D image steganography by training NeRF from scratch with the watermarked images.
    We implement two off-the-shelf steganography methods including a traditional machine learning approach called Least Significant Bit (LSB~\cite{chang2003finding}), and a deep learning pipeline as DeepStega~\cite{baluja2017hiding}.
    An ideal case is that the embedded information can be recovered from the synthesized novel views, indicating the successful NeRF steganography.
    However, as can be seen in Fig.~\ref{FIG:CHALLENGE}, the embedded information containing the hidden images cannot be recovered from the NeRF renderings.
    By analyzing the residual maps of training views (between GT training views and the watermarked) and novel views (between GT novel views and the actual rendering), 
    we observe the subtle residuals to recover the hidden information are smoothed out in NeRF renderings, and similar failures occur on other datasets as well.
    Therefore, for the rest of our study, we mainly focus on analyzing our new framework that performs direct steganography on 3D NeRF.
    

\begin{table}[t]
  \centering
  \caption{
    Quantitative results of NeRF rendering and hidden information recovery.
    We consider two conditions, embedding each scene with a common hidden image (\textit{One-for-All}) or a scene-specific hidden image (\textit{One-for-All}).
    We report the quality difference compared to the single-scene settings as  $\Delta_{SSIM}^{({10}^{-2})}$.
    Results are averaged over the selected LLFF scenes.
    }
  \resizebox{1.0\linewidth}{!}{
  \begin{tabular}{l|cccc|ccc}
    \toprule
    \multirow{2}[4]{*}{Setting} & \multicolumn{4}{c|}{NeRF Rendering} & \multicolumn{3}{c}{Hidden Recovery} \\
    \cmidrule{2-8}          & PSNR$\uparrow$   & SSIM$\uparrow$ & LPIPS $\downarrow$ & $\Delta_{SSIM}^{({10}^{-2})}$$\uparrow$   & Acc.\,(\%)$\uparrow$  & SSIM$\uparrow$  &  $\Delta_{SSIM}^{({10}^{-2})}$$\uparrow$  \\
    \midrule
    \textit{One-for-All} & \cellcolor{top2}24.99  & \cellcolor{top2}0.8013  & \cellcolor{top2}0.1786 & \cellcolor{top2}-0.10 & \cellcolor{top1}100.00 & \cellcolor{top1}0.9860  & \cellcolor{top1}+0.19 \\
    \textit{One-for-One} & \cellcolor{top1}24.99  & \cellcolor{top1}0.8016  & \cellcolor{top1}0.1779 & \cellcolor{top1}-0.07 & \cellcolor{top2}100.00 &\cellcolor{top2} 0.9122  & \cellcolor{top2}-7.19 \\
    \midrule
    \textit{One-for-All} & \cellcolor{top2}27.90  & \cellcolor{top2}0.8513  & \cellcolor{top2}0.1236 & \cellcolor{top2}+0.01 & \cellcolor{top1}100.00 & \cellcolor{top1}0.9844  & \cellcolor{top1}-0.76 \\
    \textit{One-for-One} & \cellcolor{top1}27.90  & \cellcolor{top1}0.8515  & \cellcolor{top1}0.1195 & \cellcolor{top1}+0.03 & \cellcolor{top2}100.00 & \cellcolor{top2}0.9448  & \cellcolor{top2}-4.45 \\
    \midrule
    \textit{One-for-All} & \cellcolor{top1}30.27  & \cellcolor{top1}0.8498  & \cellcolor{top1}0.1289 & \cellcolor{top1}+0.02 & \cellcolor{top1}100.00 & \cellcolor{top1}0.9430  & \cellcolor{top1}-5.42 \\
    \textit{One-for-One} & \cellcolor{top2}30.12  & \cellcolor{top2}0.8480  & \cellcolor{top2}0.1302 & \cellcolor{top2}-0.16 & \cellcolor{top2}100.00 & \cellcolor{top2}0.9102  & \cellcolor{top2}-8.67 \\
    \bottomrule
    \end{tabular}
    }
  \label{TAB:EXP-2}
\end{table}
\vspace{-1em}

\begin{figure*}[!t]   
\centering	   
\includegraphics[width=\linewidth]{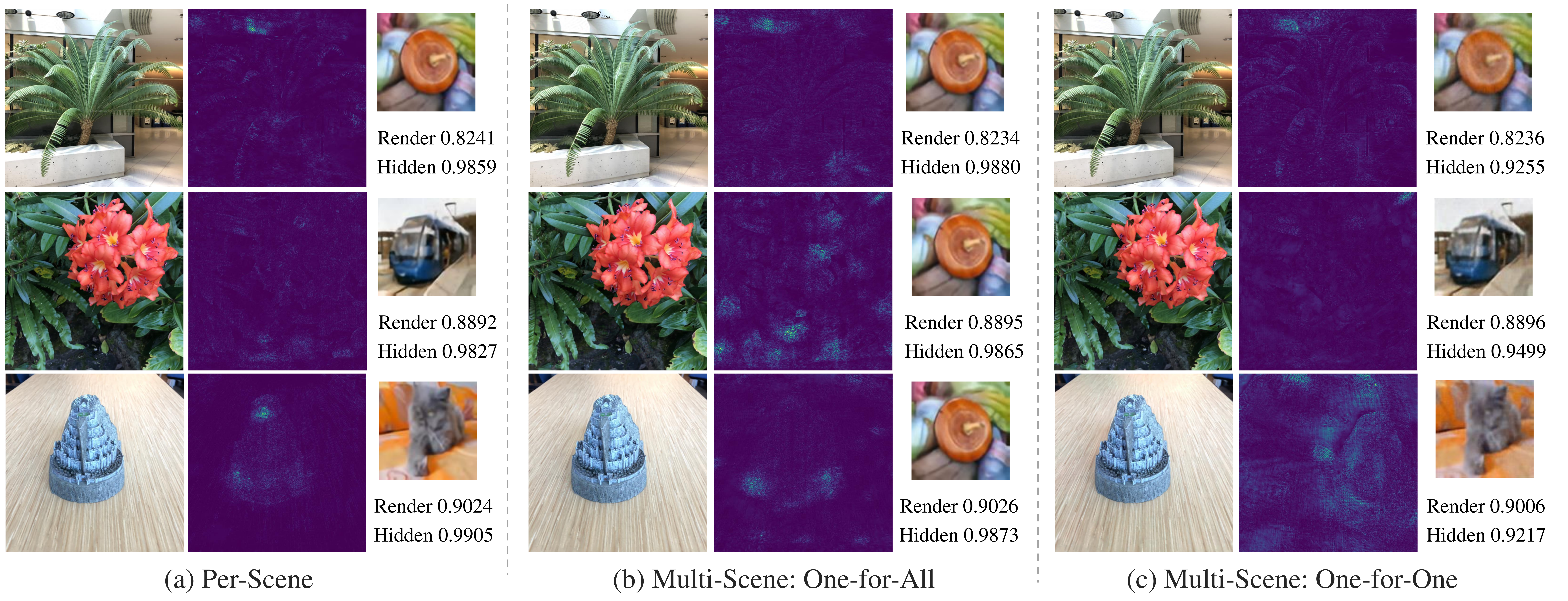}
\vspace{-1.5em}
\caption{ 
Results on three multi-scene  settings.
Within each column, we show the StegaNeRF rendering, residual error from the initial NeRF rendering, and recovered hidden image.
We show the SSIM for the StegaNeRF renderings and the recovered hidden images.
}
 \label{FIG:EXP-2}    
\end{figure*} 

    \paragraph{Results} 
    Tab.~\ref{TAB:EXP-1} contains quantitative results on LLFF and NeRF-Synthetic scenes.
    While NeRF trained by 2D steganography methods hardly recovers the embedded information, StegaNeRF accurately recovers the hidden image with minimal impact on the rendering quality measured by PSNR. 
    Fig.~\ref{FIG:EXP-1} provides qualitative results of StegaNeRF on three embedded images on the NeRF-Synthetic scenes.
    An interesting observation is the common regions where hidden information (high residual error) emerges in the renderings, \textit{e.g.}, the right rear side of \textit{lego} and left handrail of \textit{chair}.
    We also notice that, within each scene, these regions are persistent across multiple viewpoints.
    

\subsection{Case II: Embedding in Multiple Scenes.} 
    \paragraph{Settings}
    We extend our steganography scheme to embed information within  multiple scenes at once.
    Specially, we use three LLFF scenes \{\textit{flower, fern, fortress}\} to test the two sub-settings of multi-scene embedding with: (1) {\it One-for-All}, a common hidden image and (2) {\it One-for-One} scene-specific hidden images.
    All scenes share the same detector $F_{\psi}$ and classifier $F^c_{\psi}$.
    The difference between the two sub-settings is the number of hidden images that $F_{\psi}$ and $F^c_{\psi}$ need to identify and recover.
    We sample one scene for every training epoch, and due to the increased data amount, we increase training epochs until convergence.

    \paragraph{Results}
    Tab.~\ref{TAB:EXP-2} provides quantitative results on the multi-scene setting.
    %
    The performance drop compared to single-scene training is sometimes noticeable, but it is not surprising due to the inherent requirement of per-scene fitting for our NeRF framework.
    %
    Fig.~\ref{FIG:EXP-2} presents visual comparisons of multi-scene steganography against single-scene scheme.

\begin{figure*}[!t]   
\centering	   
\includegraphics[width=\linewidth]{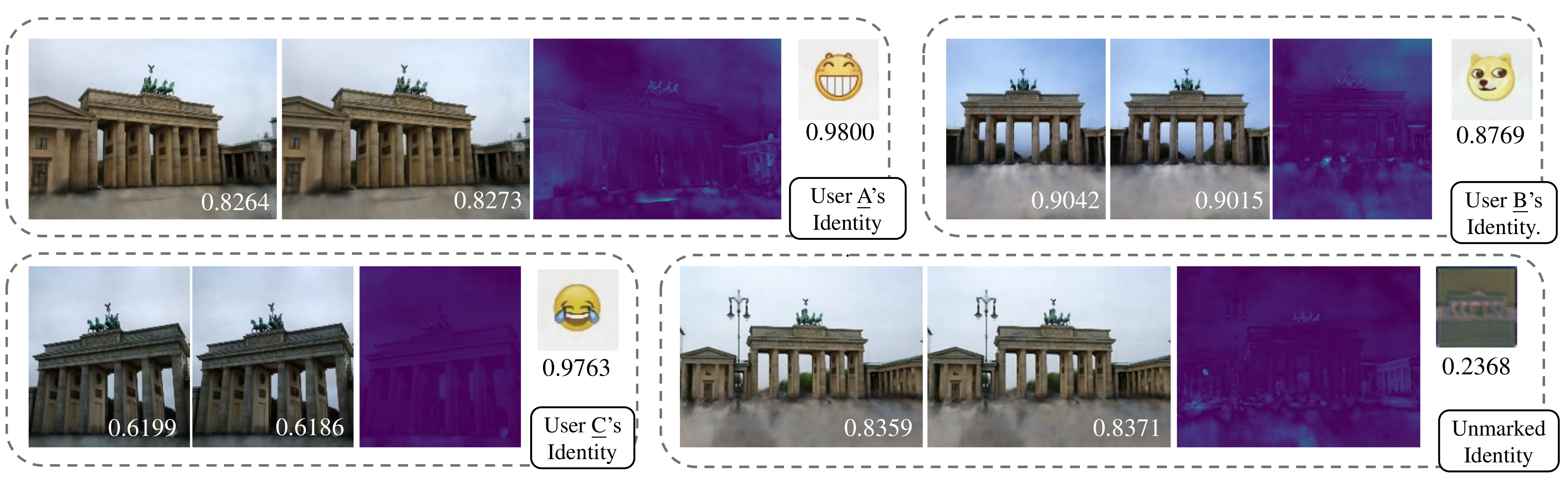}
\vspace{-1.5em}
\caption{ 
Qualitative results of NeRF steganography on NeRF-W dataset. Within each block we show the StegaNeRF rendering, the initial 
NeRF rendering, residual error between StegaNeRF and initial, and the recovered hidden image.
At bottom right, we provide the SSIM computed with their respective ground truth.
``Unmarked Identity'' denotes the remaining training views not related to any of the considered users, and the detector does not recover meaningful information from renderings at those poses.}
 \label{FIG:EXP-3}    
\end{figure*} 

\begin{figure}[!t]   
\centering	   
\includegraphics[width=\linewidth]{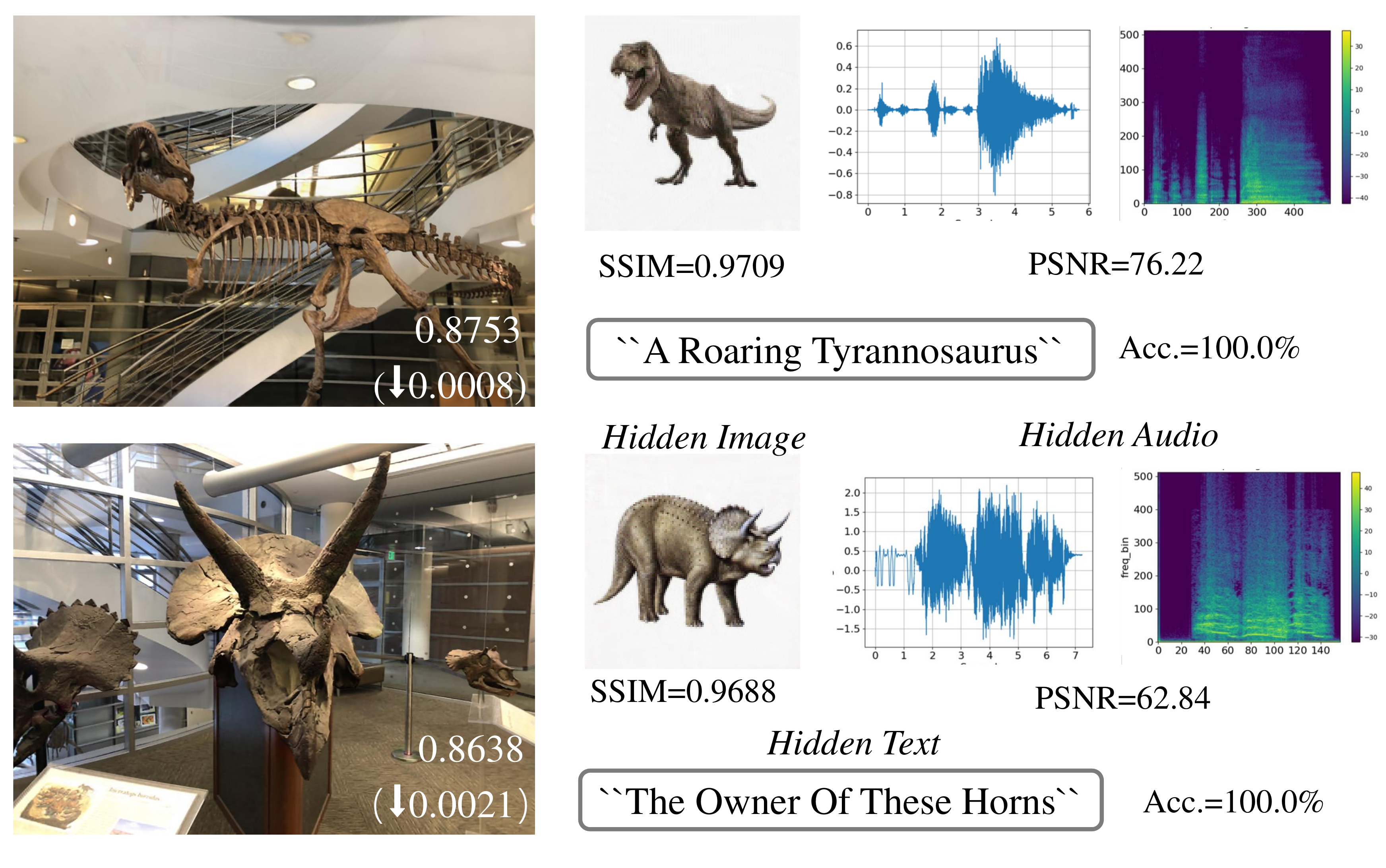}  
\vspace{-1em}
\caption{ 
Multi-modal steganographic hidden information.
Each block contains the StegaNeRF rendering, recovered hidden images, audio (shown in waveform and spectrum), and text.
The recovery metrics for hidden information in each modality are labeled respectively.
For the StegaNeRF rendering, we report both SSIM and the relative change from initial NeRF renderings.
}
 \label{FIG:EXP-4}    
\end{figure} 

\subsection{Case III: Embedding Multiple Identities.}\label{SEC:MULTI-ID}
 
\paragraph{Settings}   
Constructing NeRF of large-scale cultural landmarks is a promising application, and community contributions are crucial to form the training images.
Since every NeRF prediction is indebted to some particular training images, it would be meaningful to somehow identify the contributing users in the rendered output.
Specifically, we present a proof-of-concept based on the following scenario:
Given a collection of user-uploaded training images, 
our task is to apply StegaNeRF to ensure the final NeRF renderings hide subtle information about the relevant identities whose uploaded images help generate the current rendering.
%
%
%
%

To simulate this multi-user scenario in the public NeRF-W dataset, we randomly select $M$ anchor views with different viewing positions, and then find the $K$ nearest neighbour views to each anchor to form their respective clusters.  
We set $M=3, K=20$ in experiments.
We assume a common contributor identity for each cluster, and we want the NeRF rendering to contain customized hidden information about those $M$ contributors when we render within the spatial regions spanned by their own cluster.
Thus, our classifier network $F^{c}_{\phi}$ is modified to output $M$-class cluster predictions and another class for identity outside of the $M$ clusters. 
Since the detector should extract no information for views outside of those $M$ clusters to prevent false positives, we also compute $\mathcal{L}^-_{{dec}}$~\eqref{EQ:DEC1} and $\mathcal{L}^{c}_{dec}$~\eqref{EQ:DEC2} for those poses.


\paragraph{Results}
We employ the same network backbone as NeRF-W~\cite{tang2017automatic} to handle in-the-wild images with different time and lighting effects.
Fig.~\ref{FIG:EXP-3} presents qualitative results of embedding multiple identities in a collaborative large-scale NeRF. 

\begin{figure*}[!t]   
    \centering	   
    \includegraphics[width=\linewidth]{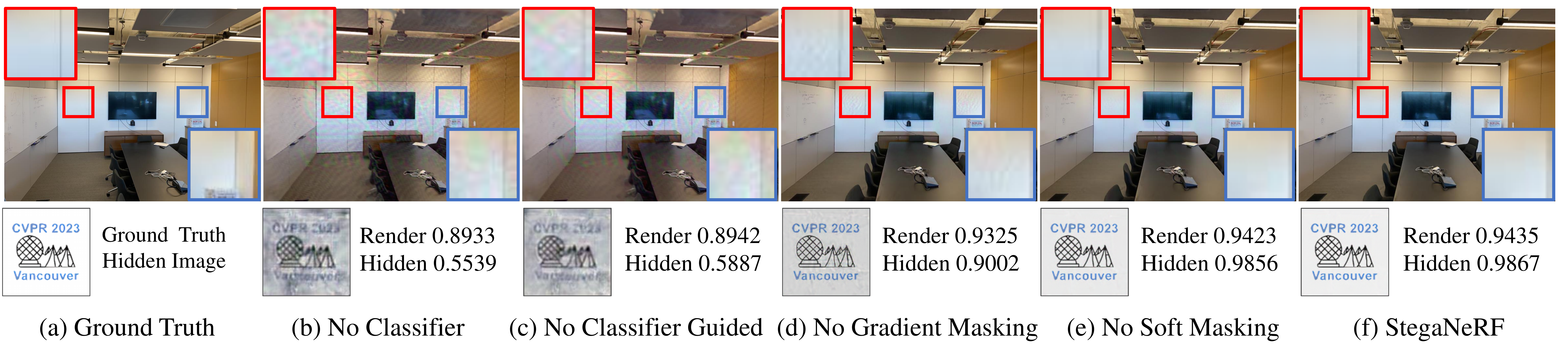}    
    \vspace{-1.5em}
    \caption{ Impact on visual quality when changing different components of the proposed StegaNeRF framework as in Tab.~\ref{TAB:EXP-5}.
    We show the SSIM for the renderings and the recovered hidden images.
    }\label{FIG:EXP-5-1}    
\end{figure*}




\subsection{Case IV: Embedding Multi-Modal Information.}

    \paragraph{Settings}
    We further show the potential of StegaNeRF in embedding multi-modal hidden information, such as images, audio, and text.
    %
    We modify the detector network to build a modal-specific detector for each modality.
    \paragraph{Results}
    Fig.~\ref{FIG:EXP-4} shows recovered multi-modal embedded signals in the \textit{trex} and \textit{horns} scenes from LLFF. 
    Evidently, the StegaNeRF framework can easily extend to embed multi-modal information with high recovery performance without scarifying the rendering quality.
    

\begin{table}[!t]
  \centering
  \caption{Ablation study of different components of StegaNeRF. Results are averaged on the selected LLFF scenes.
  }
  \resizebox{1\linewidth}{!}{
   \begin{tabular}{l|ccc|cc}
 
    \toprule
    \multicolumn{1}{l|}{\multirow{2}[4]{*}{Method}} & \multicolumn{3}{c|}{NeRF Rendering} & \multicolumn{2}{c}{Hidden Recovery} \\
\cmidrule{2-6}          & PSNR$\uparrow$  & SSIM$\uparrow$   & LPIPS$\downarrow$  & Acc.\,(\%)$\uparrow$   & SSIM$\uparrow$  \\
    \midrule
    StegaNeRF & \cellcolor{top1}28.21 & \cellcolor{top1}0.8580  & \cellcolor{top1}0.1450  & \cellcolor{top1}100.0 & \cellcolor{top1}0.9224 \\
    \midrule
    No Classifier & 26.85 & 0.8077  & 0.2417  & N/A     & 0.4417 \\
    No Classifier Guided & 27.12 & 0.8239  & 0.2073  & 100.0 & 0.5461 \\
    No Gradient Masking & \cellcolor{top3}27.86 & \cellcolor{top3}0.8375  & \cellcolor{top3}0.1710  & \cellcolor{top2}100.0 & \cellcolor{top2}0.8822 \\
    No Soft Masking & \cellcolor{top2}28.05 & \cellcolor{top2}0.8558  & \cellcolor{top2}0.1526  & \cellcolor{top3}94.44 & \cellcolor{top3}0.8751 \\
    \midrule
    Standard NeRF & 28.23 & 0.8593  & 0.1440  & 50.00 & N/A \\
    \bottomrule
    \end{tabular}%
    }
  \label{TAB:EXP-5}%
\end{table}%

\subsection{Further Empirical Analysis}
    \paragraph{Ablation Studies} 
    The effect of removing each component in the StegaNeRF framework is presented in Tab.~\ref{TAB:EXP-5}.
    %
    \textit{Standard NeRF} uses the initial results of the standard NeRF, serving as an upper-bound rendering performance.
    \textit{No Classifier} completely removes the classifier $F^c_{\psi}$, while \textit{No Classifier Guided} retains the classification task (hence impact on NeRF rendering) but does not condition the detector on classifier output.
    \textit{No Gradient Masking} removes the proposed masking strategy (Sec.~\ref{SEC:2}), and \textit{No Soft Masking} uses the binary mask with a threshold of 0.5. Our StegaNeRF makes a good balance between the rendering quality and the decoding accuracy.
    Fig.~\ref{FIG:EXP-5-1} further presents the visual impact of removing each component.

\begin{figure}[!t]  
  \centering
   \subfloat[Varying $\alpha$] 
  {
      \begin{minipage}[t]{0.5\linewidth}
          \centering          
          \includegraphics[width=\linewidth,height=2.5cm]{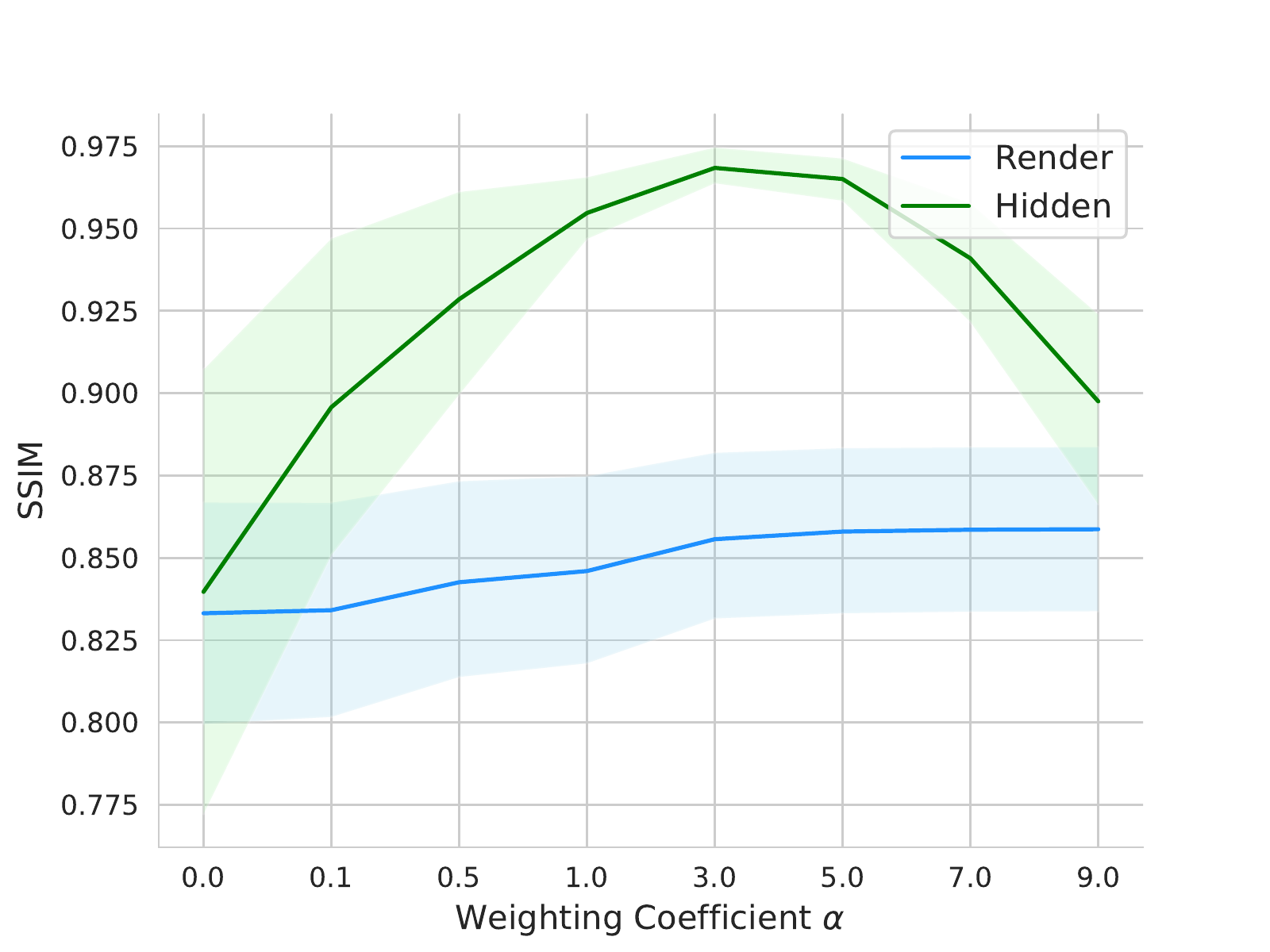} 
           \vspace{-1em}
      \end{minipage}%
 
  }%
  \subfloat[Varying masking ratio] 
  {   
       \begin{minipage}[t]{0.5\linewidth}
          \centering      
          \includegraphics[width=\linewidth,height=2.5cm]{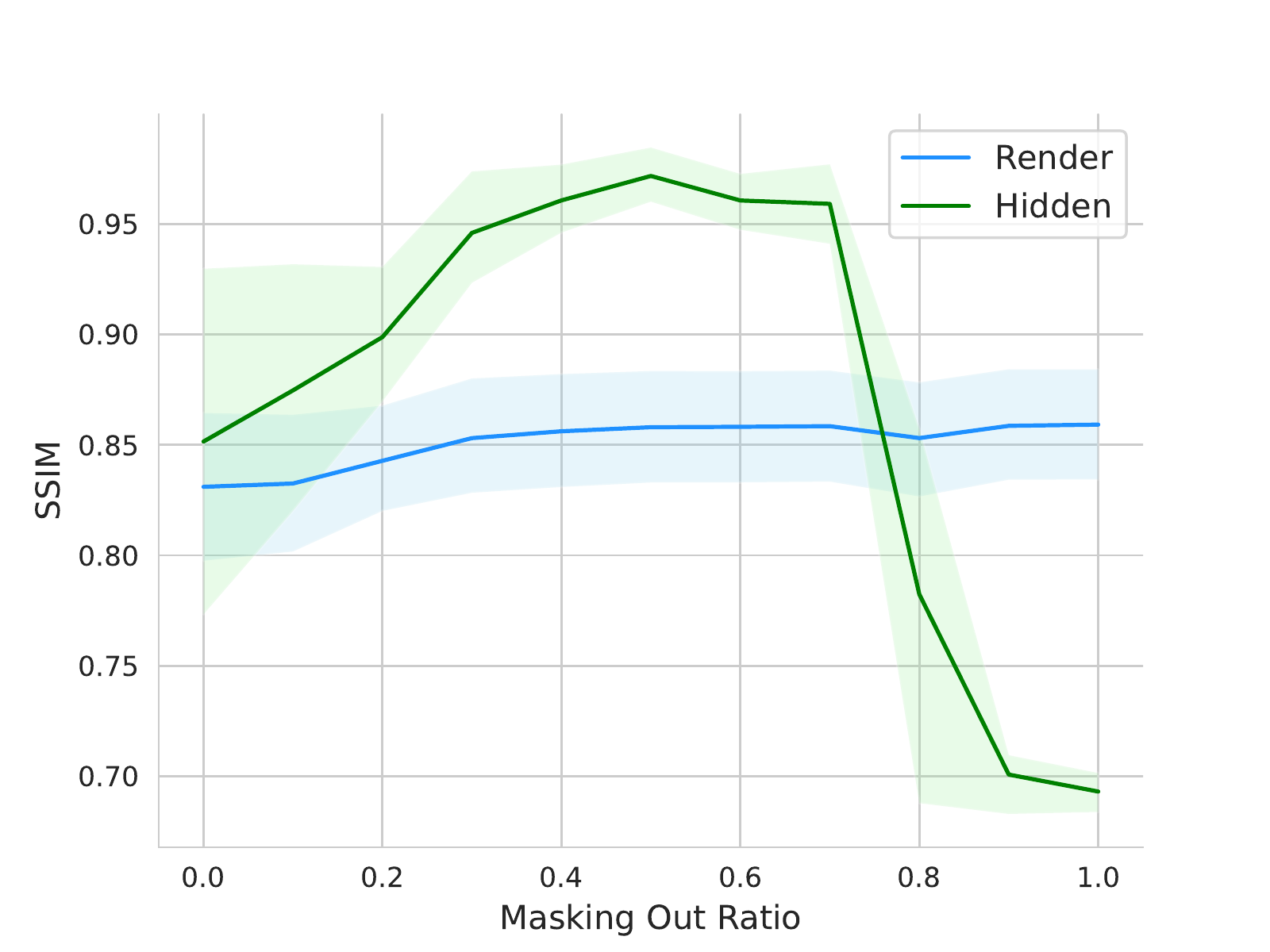} 
          \vspace{-1em}
      \end{minipage}
    }
  \caption{
  Analysis of gradient masking. 
  (a) Varying $\alpha$ from Eq.~(\ref{EQ:REG3}).
  (b) Masking different ratios of weights from the gradient updates at steganography learning.
  We provide the SSIM of rendered views (blue) and recovered hidden images (green).
  } 
   \vspace{-1em}
\label{FIG:EXP-5-2}  
\end{figure}

\begin{figure}[!t]  
  \centering
  \subfloat[Varying JPEG compression rate] 
  {
      \begin{minipage}[t]{0.5\linewidth}
          \centering      
          \includegraphics[width=\linewidth,height=2.5cm]{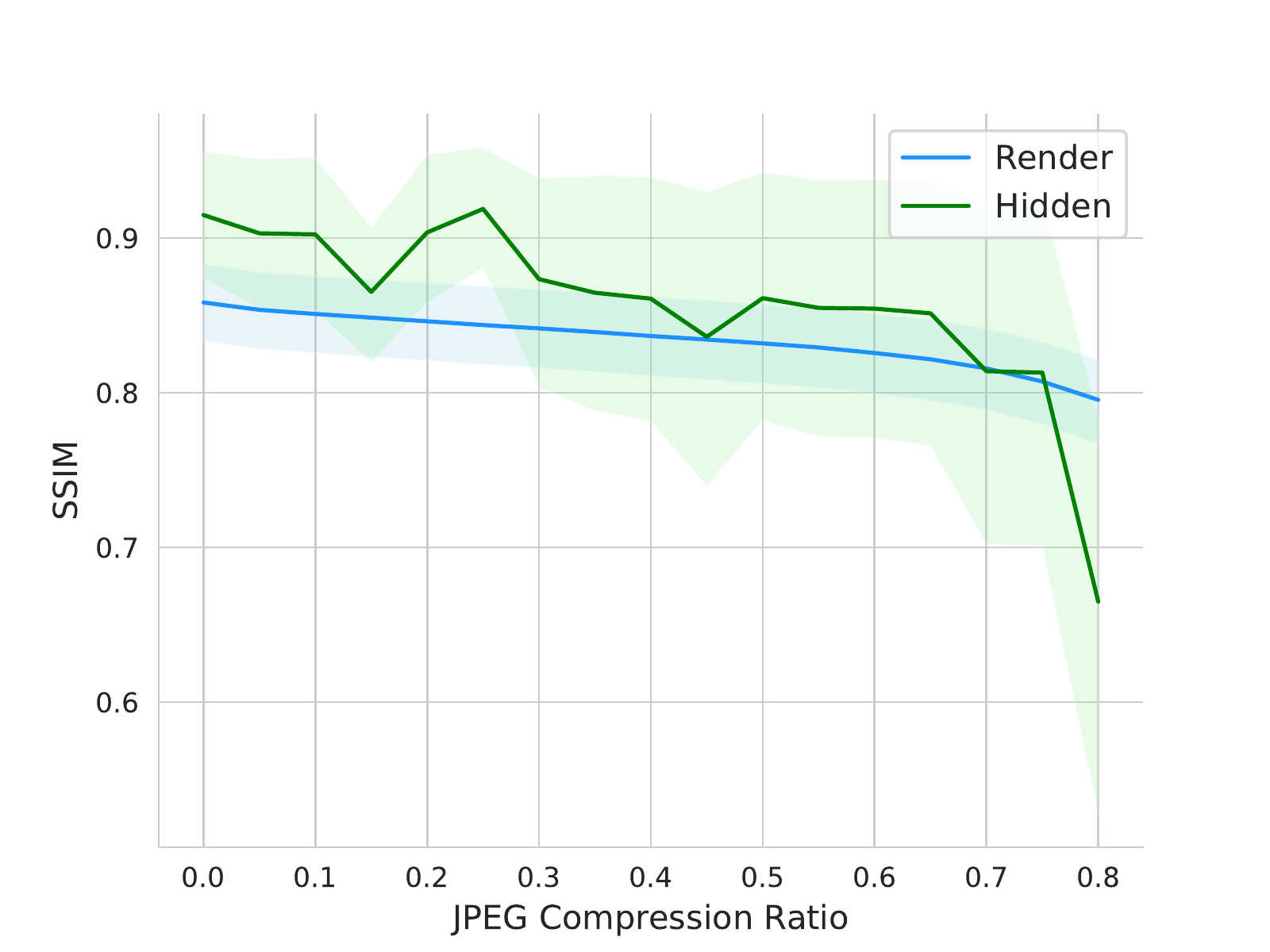} 
          \vspace{-1em}
      \end{minipage}
  }%
   \subfloat[Varying Gaussian blur std.] 
  {
      \begin{minipage}[t]{0.5\linewidth}
          \centering          
          \includegraphics[width=\linewidth,height=2.5cm]{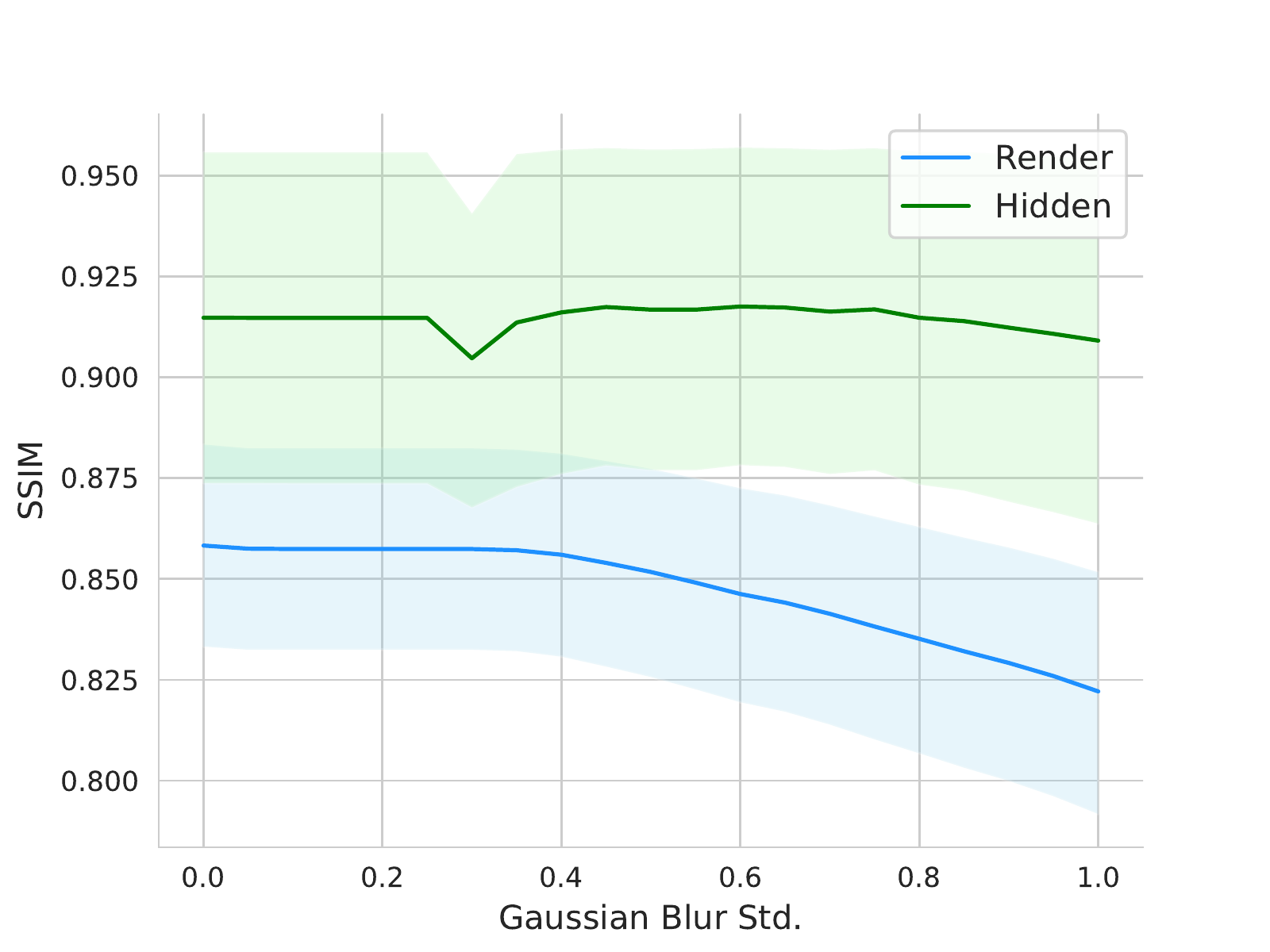} 
           \vspace{-1em}
      \end{minipage}%
  }
  \caption{
  Analysis of robustness over (a) JPEG compression and (b) Gaussian blur. 
  We provide the SSIM of rendered views (blue) and recovered hidden images (green).
  } 
\label{FIG:EXP-5-3}  
\end{figure}

\paragraph{Sensitivity Analysis}
Fig.~\ref{FIG:EXP-5-2} shows the impact of varying gradient masking strategies.
Fig.~\ref{FIG:EXP-5-3} reports the performance of StegaNeRF against the common perturbations including JPEG compression and Gaussian noise.
The lines show the average accuracies across selected scenes and the shaded regions indicate the range of 0.5 standard deviation.
{The hidden recovery of StegaNeRF is robust to various JPEG compression ratios and Gaussian blur degradation.
}


\section{Discussion}
\label{SEC:DISCUSSION}
In this section, we provide further discussion on the significance of the proposed framework and presented results.

    %

\paragraph{Why 2D Steganography fails in NeRF?}
The changes to the NeRF training images induced by 2D steganographic methods are hard to be retained in NeRF renderings.
This is not surprising as NeRF tends to smooth out the subtle details from training images.
In contrast, our method directly optimizes the NeRF weights so that its rendering contains subtle details that the detector network can identify.
{See the supplement for a more detailed analysis.}

\paragraph{How useful is steganography for NeRF?} 
Although NeRF-based 3D content has yet to become mainstream, we believe it will play a major future role not only for social platforms, but also for 3D vision research and applications.
On the one hand, when people upload their personal NeRF models online for viewing purposes, NeRF steganography for ownership identification is apparently an important feature.
On the other hand, future 3D vision research will likely demand large-scale datasets with NeRF models trained with real-world images, and in this context, NeRF steganography can be a crucial tool for responsible and ethical uses of training data and deployed models.

\paragraph{Why not just share the rendered images instead of NeRF weights?}
Directly sharing the NeRF model gives end users the freedom to render at continuous viewpoints, which fundamentally differs from the traditional setting centered around discrete 2D image frames.
It would not be realistic to assume people will stick with sharing 2D images instead of 3D weights, and therefore, it is necessary for the research community to see beyond 2D steganography and explore 3D NeRF steganography.

\paragraph{What are the limitations?} 
The current implementation of StegaNeRF is limited by the computational efficiency of NeRF, since the steganographic training stage requires performing gradient descent updates through the NeRF model. We believe future work can further reduce the training time of NeRF and alleviate this issue.
Another limitation is that, the recovery accuracy of the injected hidden information varies across different 3D scenes.
More systematic analysis is required to understand the underlying factors that make some scenes easier to hide information than others.
%
    




\section{Conclusion}
As NeRF-based content evolves into an integral component of various 3D applications, it becomes imperative for the research community to address the upcoming demand and necessity to embed information in NeRF models as they are exchanged and distributed online.
This work presents StegaNeRF, the first framework of automatic steganographic information embedding in NeRF renderings.
Our study details the relevant components required to achieve the desired functionality, revealing potential pitfalls and insights useful for future developers.
This paper offers a promising outlook on ownership identification in NeRF and calls for more attention and effort on related problems.

\normalem
{\small
\bibliographystyle{ieee_fullname}
\bibliography{bib}
}

\end{document}